\documentclass[runningheads]{llncs}
\usepackage{comment}
\usepackage{amsmath,amssymb} % define this before the line numbering.
\usepackage{color}
\usepackage{epsfig}
\usepackage{graphicx}
\usepackage{amsmath}
\usepackage{amssymb}
\usepackage{graphicx} % DO NOT CHANGE THIS
\usepackage{amsfonts}       % blackboard math symbols
\usepackage{booktabs}       % professional-quality tables
\usepackage{floatrow}
\usepackage{multirow, hhline}
\usepackage{bm}
\newfloatcommand{capbtabbox}{table}[][\FBwidth]

\newcommand{\thetab}{\bm{\theta}}
\newcommand{\betab}{\bm{\beta}}
\newcommand{\mub}{\bm{\mu}}

\newcommand{\bphi}{\bm{\phi}}

\newcommand{\tb}{\mathbf{t}}
\newcommand{\vb}{\mathbf{v}}

\newcommand{\zz}{\mathbf{z}}

\newcommand{\xx}{\mathbf{x}}
\newcommand{\yy}{\mathbf{y}}
\newcommand{\ff}{\mathbf{f}}
\newcommand{\hh}{\mathbf{h}}
\newcommand{\eg}{e.g.\ }
\newcommand{\ie}{i.e.\ }
\newcommand{\Ib}{\mathbf{I}}

\newcommand{\Sb}{\mathbf{S}}

\newcommand{\Vb}{\mathbf{V}}

\DeclareMathOperator*{\argmin}{arg\,min}

\begin{document}
\pagestyle{headings}
\mainmatter
\def\ECCVSubNumber{6296}  % Insert your submission number here

\title{Weakly Supervised 3D Human Pose and Shape Reconstruction with Normalizing Flows} % Replace with your title

\begin{comment}
\titlerunning{ECCV-20 submission ID \ECCVSubNumber} 
\authorrunning{ECCV-20 submission ID \ECCVSubNumber} 
\author{Anonymous ECCV submission}
\institute{Paper ID \ECCVSubNumber}
\end{comment}
%******************

% \begin{comment}
\titlerunning{Weakly Supervised 3D Human Pose and Shape with Normalizing Flows}
\author{Andrei Zanfir \and
Eduard Gabriel Bazavan \and
Hongyi Xu \\
William T. Freeman \and
Rahul Sukthankar \and
Cristian Sminchisescu}
\authorrunning{A. Zanfir et al.}
\institute{Google Research
\email{\{andreiz,egbazavan,hongyixu,wfreeman,sukthankar,sminchisescu\}@google.com}}
% \end{comment}
%******************
\maketitle

\begin{abstract}
Monocular 3D human pose and shape estimation is challenging due to the many degrees of freedom of the human body and the difficulty to acquire training data for large-scale supervised learning in complex visual scenes.
In this paper we present practical semi-supervised and self-supervised models that support training and good generalization in real-world images and video. 
Our formulation is based on kinematic latent normalizing flow representations and dynamics, as well as differentiable, semantic body part alignment loss functions that support self-supervised learning. 
In extensive experiments using 3D motion capture datasets like CMU, Human3.6M, 3DPW, or AMASS, as well as image repositories like COCO, we show that the proposed methods outperform the state of the art, supporting the practical construction of an accurate family of models based on large-scale training with diverse and incompletely labeled image and video data.

\end{abstract}

\section{Introduction}

Recovering 3D human pose and shape from monocular RGB images is important for motion and behavioral analysis, robotics, self-driving cars, computer graphics, and the gaming industry. Considerable progress has been made recently in increasing the size of datasets, in the level of detail of human body modeling, and the use of deep learning. A difficulty is the somewhat limited diversity of supervision available in the 3D domain. Many datasets offer 2D human body joint annotations or semantic body part segmentation masks for images collected in the wild, but lack 3D annotations. Motion capture datasets in turn offer large and diverse 3D annotations but their image backgrounds, clothing or body shape variation is not as high. Multi-task models, or models able to learn using limited forms of supervision, represent a potential solution to the current 3D supervision limitations. However, the number of human body shapes and poses observed in images collected in the wild is large, so strong pose, shape priors and expressive loss functions appear necessary in order to make learning feasible. In this paper we address some of these challenges by designing a family of normalizing flow based kinematic priors, together with semantic alignment losses that make large scale weakly and self-supervised learning more accurate and efficient. \emph{The introduction and integration of these components, new in the framework of human sensing, with strong results, is one of the main contributions of this work}. An evaluation (with ablation studies) on large scale datasets like Human3.6M, COCO, 3DPW, indicates good weakly supervised performance for 3D reconstruction. Our proposed priors and loss functions are amenable to both integration into deep learning losses and to direct non-linear state optimization (refinement) of a model given a random seed or initialization from a learnt predictor. \\
\noindent{\bf Mindset.} Our use of different data sources is practically minded, as we aim towards large scale operation in the wild. Hence we rely on all types of supervision and data sources available.
We often start with models trained in the lab, \eg using Human3.6M and those are \emph{supervised}. We also use 3D motion capture repositories like CMU in order to construct kinematic (output) priors and that component alone would make our approach \emph{semi-supervised}. Finally, we make use of large scale predict-and-reproject losses for unlabeled datasets like MS COCO, which makes our approach, at least to an extent \emph{self-supervised}. Whatever model curriculum used, we aim, long-term, to converge on self-supervised operation. We work with a semi-supervised output prior and model ignition is based on supervision in the lab. By convention, we call this regime \emph{weakly-supervised}.

\noindent{\bf Related Work.} There is considerable work in 3D human pose estimation based on 2D keypoints, semantic segmentation of body parts, and 3D joint positions \cite{rogez2016mocap,Fua17,mehta2017vnect,dmhs_cvpr17,martinez17iccv,iskakov2019learnable}. More recently, there has been significant interest in 3D human pose and shape estimation \cite{sun2019human,doersch2019sim2real,kanazawa2019learning,kolotouros2019convolutional,arnab2019exploiting,xu2019denserac}, with some in the form of a reduced parametric model \cite{pavlakos2017cvpr} decoded by 2D predictions, volumetric variants \cite{varol18_bodynet} or direct vertex prediction combined with 3D model fitting \cite{jackson20183d,zanfir2018monocular}. Learning under weak supervision represents the next frontier, considered in this work as well. \cite{yang20183d} learns a discriminator in order to transfer knowledge gained on a 3D dataset to a 2D one. \cite{zhou2017towards} train a shared representation for both 2D and 3D pose estimation, with a regularizer operating on body segments in order to preserve statistics.
\cite{Kanazawa2018} use a discriminator as prior, with adversarial training, and mixes 3D supervision and image labels. \cite{omran2018nbf} uses segmentations as an intermediate layer, defines a loss on 2D and 3D joints, and rely on rotation matrices instead of angle-axiss representation. 
\cite{pavlakos2018learning} uses a differentiable renderer (OpenDR) to compute a silhouette loss with a limited basin of attraction. This is only used for finetuning the network, but the authors report not having observed significant gains. \cite{NIPS2017_7108} rely on a segmentation loss defined on silhouettes, not on the body parts, and rely on multiple views and temporal constraints for learning. 

A variety of methods rely on priors for 3D optimization starting from an initial estimate provided by a neural network and/or by relying on image features like keypoints or silhouettes.
\cite{bogo2016} fit a Gaussian mixture model to motion capture data from CMU \cite{cmu2018} and use it during optimization. We will evaluate this prior in our work. SPIN\cite{kolotouros2019learning} alternates rounds of training with estimation of new targets using optimization (we will compare in \S\ref{sec:exps}).

Multiple differentiable rendering models \cite{kato2018neural,loper2014opendr,rhodin2015versatile} have been proposed recently, in more general settings. Such models are elegant and offer the promise of optimizing photo-realistic losses in the long run. The challenge is in defining an end-to-end model that embeds the difficult assignment problem between the model predictions (rasterized or not) and the image features in ways that are both differentiable and amenable to larger basins of attraction. Our semantic alignment loss is not technically a rendering model, but is differentiable and offers large basins of attraction, supported by explicit, long-range semantic body part correspondences. Gradients can be propagated for points that are not rendered (\ie points that fail the z-test) and the operation is parallelizable and easy to implement.

\section{Methodology}

\noindent{\bf 3D Pose and Shape Representations.}
We use a statistical body model \cite{SMPL2015,xu2020ghum} to represent the pose and the shape of the human body. Given a monocular RGB image, our objective is to infer the pose state variables $\thetab \in \mathbb{R}^{N_{j}\times3}$ and shape $\betab \in \mathbb{R}^{N_{s}}$. A posed mesh $\mathbf{M}(\thetab, \betab)$ has $N_{v}$ associated 3D vertices $\mathbf{V}=\{\mathbf{v}_{i}, i=1\ldots N_v\}$. By dropping dependency on parameters we sometimes denote $\mathbf{M}(\mathbf{V}, k)$ the subset of vertices associated with body part index $k$ (\eg torso or head). 
%\ldots, \mathbf{v}_{N_{v}} ]\right.$ 
%\ldots, \mathbf{v}_{N_{v}} ]\right.$ 

For prediction and optimization tasks we experiment with several kinematic representations. The angle-axis gives good results in connection with deep learning architectures \cite{Kanazawa2018,zanfir2018monocular}. The representation consists of a set of $N_{j}$ angle-axis variables $\thetab = \{\thetab_{1}, \thetab_{2}, \ldots, \thetab_{N_{j}}\}, \thetab_{i} \in \mathbb{R}^{3}$, where the norm of $\thetab_{i}$ is the rotation angle in radians and $\frac{\thetab_{i}}{\left\| \thetab_{i}\right\|}$ is the unit length 3D axis of rotation. 

We also explore a new 6D over-parameterization of rotations \cite{zhou2018continuity}, given by the first two columns of the rotation matrix.
We test this parameterization in the context of optimization, by building a prior and minimizing a cost function over the compound space of 6D kinematic rotations.\footnote{We have also considered quaternions, but our experiments showed these to be inferior even to  angle-axis (AA), by at least 10\%.}

\subsection{3D Normalizing Flow-based Representations}

\noindent{\bf Existing Work on 3D Human Priors.} The method of \cite{bogo2016} builds a density model to favor more probable poses over improbable ones. They use a mixture with 8 Gaussian modes $N(\mub_j,\bm{\Sigma}_j)$, fitted to 1 million CMU poses. During optimization, the prior is evaluated to produce the log-likelihood of the pose. For numerical stability and to avoid excessive averaging effects, an approximation based on choosing the closest mode is used, which is not smooth, and may still lead to instability during mode switching. 

For neural network models, \cite{Kanazawa2018} proposed a factorized adversarial network to learn the admissible rotation manifold of 3D poses, by relying on $N_{j}+1$ discriminators, one for each joint, and one for the whole pose. The rotation limits for each joint are expected to be learned implicitly by each of the $N_{j}$ discriminators, while the last one measures the probability of the combined pose. Learning rotation matrices (as opposed to angle-axis based) discriminators, is beneficial in avoiding the non-continuous nature of the angle-axis representation, but trades off increasing representational redundancy and consequently dimensionality.

Another approach has been pursued by \cite{pavlakoscvpr2019}, where the authors use a variational auto-encoder for 3D poses. The reconstruction loss is the mean per-vertex error between the input posed mesh and the reconstructed one. The latent representation can be used as a prior, by querying the log-likelihood of a given pose.  Our experiments with VAEs constructed on top of kinematic representations (joint angles, rotations) showed that those have poor performance compared to our proposed models. The more sophisticated approaches used in VPoser \cite{pavlakoscvpr2019} rely on losses defined on meshes rather than kinematics, but meshes inevitably introduce artefacts due to \eg imperfect skinning.
Moreover, VAEs need to balance two terms -- the reconstruction loss and a KL divergence, which leads to a compromise: either the latent space is not close to Gaussian or/and decoding is imperfect. Our normalizing flow approach ensures that reconstruction loss is perfect (by the bijectivity of NFlow's construction) and during training we only optimize against the simpler Gaussian latent space objective. 

 \noindent{\bf Normalizing Flow Priors.} In this paper we propose different normalizing flow-based prior representations, to our knowledge used for the first time in modeling 3D human pose. A normalizing flow \cite{rezende2015variational,dinh2014nice,dinh2016density,kingma2018glow} is a sequence of invertible transformations applied to the original distribution. The end-result is a warped (latent) space with a potentially simple and tractable density function, \eg $\zz \sim \mathcal{N}(0; \Ib)$). We consider $\thetab \sim p^*(\thetab)$ sampled from an unknown distribution. One way to learn it is to use a dataset $\mathcal{D}$ (\eg from CMU or Human3.6M) and maximize data log-likelihood with respect to a parametric model $p_{\bphi}(\thetab)$
\begin{align}
    \max_{\bphi} \sum_{\thetab \in \mathcal{D}} \log p_{\bphi}(\thetab)
\end{align}
where $\bphi$ are the parameters of the generative model.
If we choose $\zz = \ff_{\bphi}(\thetab)$ where $\ff_{\bphi}$ is a component-wise invertible transformation, one can rewrite the log-probability under a change of variables
\begin{align}
    \log p_{\bphi}(\thetab) = \log p_{\bphi}(\zz) + \log \left|\det(d\zz/d\thetab)\right| \label{eq:nflow_change}
\end{align}
Dropping the subscript $\bphi$, if $\ff$ is the composition of multiple bijections $\ff_i$, with intermediate output $\hh_i$, \eqref{eq:nflow_change} becomes
\begin{align}
    \log p_{\bphi}(\thetab) &= \log p_{\bphi}(\zz) + \sum_{i=1}^{K} \log \left|\det(d\hh_i/d\hh_{i-1})\right| \label{eq:nflow_change_factorized} 
\end{align}
where $\hh_0 = \thetab$ and $\hh_K = \zz$, and $p_{\bphi}(\zz) = \mathcal{N}(\zz; \mathbf{0}, \Ib)$ is chosen as a spherical multivariate Gaussian distribution. State-of-the-art flow architectures are based on auto-regressive versions, such as the Masked Autoregressive Flow (MAF) \cite{papamakarios2017masked}, Inverse Autoregressive Flow (IAF) \cite{kingma2016improved}, NICE \cite{dinh2014nice}, MADE \cite{germain2015made} or Real-NVP \cite{dinh2016density}. In our experiments, we found MAF/IAF/MADE to be too slow given our representation and dataset size, with no measurable improvement over a Real-NVP. A Real-NVP step takes as input a variable $\xx$ and outputs the transformed variable $\yy$, under the following rules
\begin{align}
    \yy_{1:d} = \xx_{1:d}, \quad
    \yy_{d+1:D} = \xx_{d+1:D} \odot \exp{\mathbf{s}} + \tb,
\end{align}
where $\mathbf{s}$ and $\tb$ are shift-and-scale vectors that can be modelled as neural network outputs,
\ie $(\mathbf{s}, \tb) = \text{NN}(\xx_{1:d})$, and $d$ is the splitting location of the current $D$-dimensional variable. The '$\odot$' operator represents the pointwise product, while '$\exp$' is the exponential function. In order to chain multiple Real-NVP steps, one has to ensure that order is not constant, otherwise the first $d$-dimensions would not be transformed. Typically, $\xx$ is permuted before the operation. Because, in our case, $\thetab$ has moderate although sufficiently large size, we introduce a trainable, fully-connected layer before each NVP step. This is fast and results in better models. We also experiment with a lower capacity model, which replaces the Real-NVP with a simple parametric ReLU, as activation function. We do not use batch normalization. We found that we can trade a bit of accuracy (given by RealNVP) for a standard MLP that is faster and requires less memory.  For the same network depth, the Real-NVP variant had 2x the number of parameters, and had marginal performance benefits (2\%). More details can be found in the Sup. Mat.

For optimization-based inference or neural network training, we can parameterize the problem either in the latent (warped) space, or in the ambient (original) kinematic space, given the exact connection between them. Our empirical studies show that directly predicting (or optimizing) the latent representation always yields better results over working in the ambient space (see table \ref{tbl:fitting_experiment_h80k}).

\begin{figure}[!ht]
\begin{center}
    \includegraphics[width=0.8\linewidth]{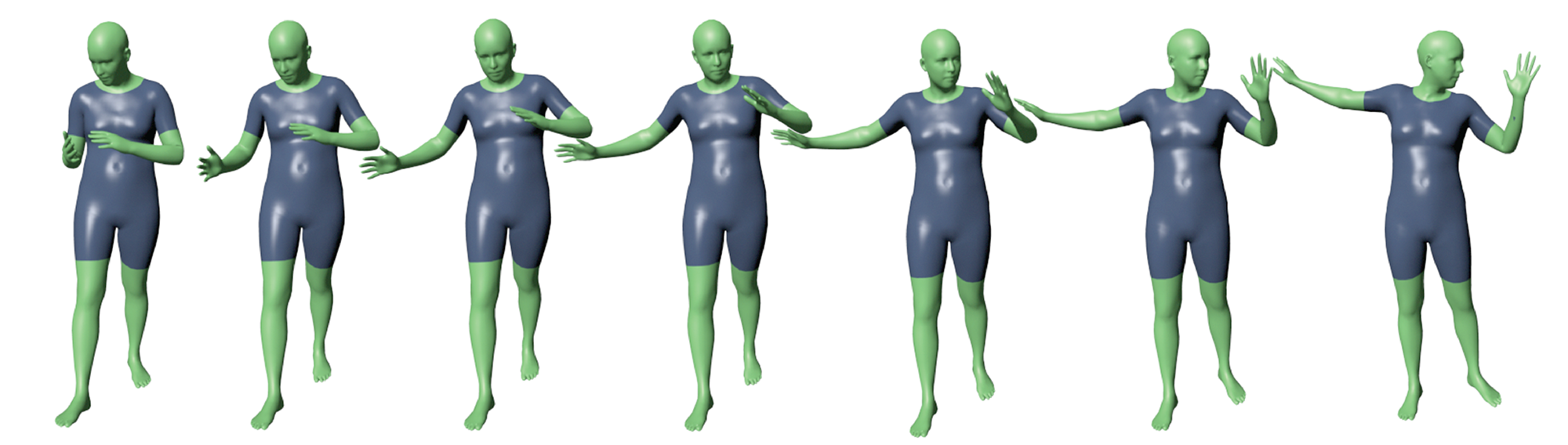}
\end{center}
\caption{\small From left to right: interpolation in latent space for normalizing flow, for two (begin and end) normal random codes. Notice smooth results, plausible human poses.}
\label{fig:interpolation}
\end{figure}

In fig. \ref{fig:interpolation} we show a sample pose interpolation in latent space.

\noindent{\bf Optimization.} To optimize normalizing flow representations, we assume normalization variable $\thetab$, Gaussian variable $\zz = \ff(\thetab)$, and $\thetab = \ff^{-1}(\zz)$, given that $\ff$ is bijective. We define the normalizing flow prior as the negative log-likelihood in ambient $\psi_{nf}(\ff(\thetab)) = -\log p_{\bphi}(\ff(\thetab))$ or, equivalently in latent space, $\psi_{nf}(\zz) = -\log p_{\bphi}(\zz)$. Then, for any objective function or loss defined as $L(\thetab, \betab)$, we have either the option of working (\ie predicting or optimizing) in the \textbf{ambient space} and back-projecting in the latent space at each step
\begin{align}
    \argmin_{\thetab, \betab} L(\thetab, \betab) + \psi_{nf}(\ff(\thetab)) \label{eq:nflow_orig}
\end{align}
or the option to operate in the \textbf{latent space} directly
\begin{align}
    \argmin_{\zz, \betab} L(\ff^{-1}(\zz), \betab) +  \psi_{nf}(\zz) \label{eq:nflow_wrp}
\end{align}
Both approaches are differentiable and we will evaluate them in \S\ref{sec:exps}.

\subsection{Differentiable Semantic Alignment Loss}

In order to be able to efficiently learn using weak supervision (\eg just images of people), one needs a measure of prediction quality during the different phases of model training. 
In this work we explore forms of structured feedback by considering detailed correspondences between the different body part vertices of our 3D human body mesh (projected in the image), and the semantic human body part segmentation produced by another neural network. 

As presented by \cite{zanfir2018monocular}, an Iterated Closest Point (ICP)-style cost for body part alignment can be designed in 2D (for 3D this is quite common \eg  \cite{zhang2017detailed}). Given a set of $N_b$ body parts, their semantic image segmentation $\{\Sb_i \subset \mathbb{R}^2\}$ and associated mesh vertices of similar type $\{\mathbf{M}(\Vb, k) \subset \mathbb{R}^3\}$ (\ie the 3D vertex set of body part $k$), a distance between the set of semantic segmentation regions and the 3D mesh vertex projections (using an operator $\Pi$) can be defined as the first term of \eqref{eq:ba_loss}.
This term encourages pixels of a particular semantic body type (\eg torso, head or left lower arm) to attract projected model vertices with the same body part label. Depending on the sizes of the image regions with particular labels, and the corresponding number of vertices, the minimum of this function is not necessarily achieved only when all vertices are inside the body part. Consequently, we add a complementary loss, encouraging good overlap between  model projections and image regions of corresponding semantics 
\begin{align}
    L_{BA}(\Sb, \Vb) =& \sum_{k=1}^{N_b} \sum_{\mathbf{p} \in \mathbf{S}_k} \min_{\vb \in \mathbf{M}(\mathbf{V},k)} \left\lVert \mathbf{p} - \Pi(\vb)\right\rVert + 
  %  \\
%    & 
    \sum_{k=1}^{N_b}\sum_{\vb \in \mathbf{M}(\mathbf{V},k)} \min_{\mathbf{p} \in \Sb_k} \lVert \mathbf{p} - \Pi(\vb)\rVert \label{eq:ba_loss}
\end{align}
We will refer to the two terms as the forward semantic segmentation loss and the backward loss, respectively. Compared to state-of-the-art differentiable rendering techniques like \cite{kato2018neural}, this loss has exact gradients, because we express it as an explicit objective connecting semantic image masks and mesh vertex projections. Furthermore, our method is designed for categorical masks and only defined for regions explained by the vertex projections of our model, rather than all the image pixels.
The process is naturally parallelizable, and we offer a GPU implementation.

\subsection{Network Architecture}

\begin{figure}[h!]
\begin{center}
    \includegraphics[width=0.8\linewidth]{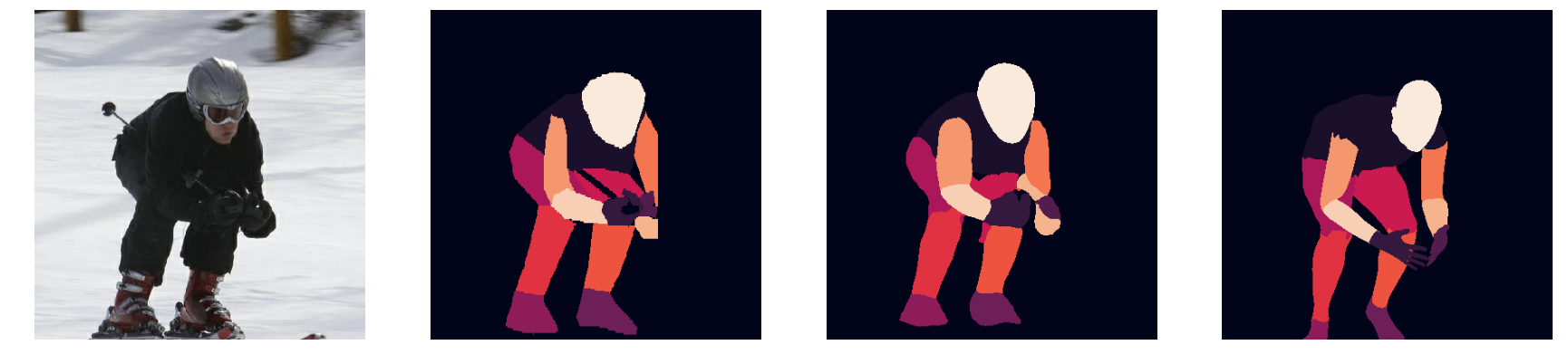}
\end{center}
\caption{\small From left to right: Original image, ground truth semantic body part segmentation mask from MSCOCO 2014, predicted segmentation mask, projected semantic mask of our 3D mesh.}
\label{fig:segmentation}
\end{figure}

Our architecture is based on a multistage deep convolutional neural network to predict human body joints, semantic segmentation of body parts, as well as 3D body pose and shape. The network consists of multiple modules, and has multiple losses, each corresponding to a different prediction task, but it can be run with a subset of the losses under different levels of supervision ranging from full to none. The first module takes as input the image and outputs keypoint (body joint) heatmaps \cite{Cao2017paf2d}. We extract the joint positions from the heatmaps and obtain $\mathbf{J}_{2d} = \{ \mathbf{J}_{i}, i=1 \ldots N_{j} \}$. The next module computes semantic body part segmentations by processing images and the keypoint heatmaps obtained by the keypoint prediction module. The outputs are semantic segmentation heatmaps for each body part (see fig. \ref{fig:segmentation}), $\mathbf{S} = \{\mathbf{S}_{i}, i=1 \ldots N_{b} \}$. The last module predicts pose and shape parameters. It takes as input the outputs from previous modules and produces $\{\thetab,\betab\}$. For the camera, we adopt a perspective projection model. We fix the intrinsics and estimate translation by means of fitting the predicted 3D skeleton to 2D joint detections (that step alone requires a weak perspective approximation, see Sup. Mat.).

\noindent{\bf 3D Pose and Shape.}
The goal of the 3D pose layers is to predict the pose and shape parameters $\{\thetab, \betab\}$. The associated network is similar to the ones of the previous two modules. A stack of convolutional stages is created with losses on each stage to reinforce the weights and avoid vanishing gradients  \cite{wei2016cpm,Cao2017paf2d}. The architecture of each 3D regressor stage is composed of a stack of 5 x 2D convolutional layers with 128 feature maps, 7x7 kernels, \verb|relu| activations, followed by another 2D convolutional module with 128 layers and  1x1 kernels. The last layer is a 2D convolutional layer, has no activation function and the number of heatmaps is equal to the number of predicted parameters. Two separate dense layers are used to output $\{\thetab, \betab\}$.

\subsubsection{Supervised and Weakly Supervised Losses}\label{sec:loses}
We train our network by using a combination of fully and weakly supervised losses. The fully supervised training regime assumes complete ground truth on pose, shape. A predicted posed mesh $\mathbf{M}(\thetab, \betab)$ with $N_{v}$ associated vertices $\mathbf{V}=\{\mathbf{v}_{i},i=1\ldots N_v\}$ 
has ground truth $\mathbf{M}(\widehat{\thetab}, \widehat{\betab})$ with vertices $\{\widehat{\mathbf{v}}_{i}\}$.
We define the following MSE losses, respectively, on the mesh 
\begin{equation}
L_{V} = \frac{1}{N_{v}} \sum_{i=1}^{N_v}\left\|\mathbf{v}_{i} - \widehat{\mathbf{v}}_{i}\right\|_{2}^{2}
\end{equation}
pose and shape parameters
\begin{equation}
L_{\thetab} = \frac{1}{N_{j}} \sum_{i=1}^{N_j}\left\|\thetab_{i} - \widehat{\thetab}_{i}\right\|_{2}^{2}, \;\;\;
L_{\betab} = \frac{1}{N_{s}} \sum_{i=1}^{N_{s}}\left\|\betab_{i} - \widehat{\betab}_{i}\right\|_{2}^{2}
\end{equation}
The supervised loss combines previously defined losses
\begin{equation}
L_{fs} = L_{V} + L_{\thetab} + L_{\betab}
\end{equation}
For the weakly supervised case, the predicted mesh $\mathbf{M}(\thetab, \betab)$ is projected into the image. Denote the projected skeleton joints by $\mathbf{J}_{2d} = \{\mathbf{J}_i\}$,
the estimated (or ground truth) 2D joint positions by $\widehat{\mathbf{J}}_{2d} = \{\widehat{\mathbf{J}}_i\}$, and the semantic body part segmentation maps by $\widehat{\mathbf{S}}=\{\widehat{\mathbf{S}}_i,i=1\ldots N_{b}\}$. The weakly supervised regime assumes access to large 3D mocap datasets, \eg CMU -- in order to construct kinematic priors -- but without the corresponding images. Additionally we also rely on images in the wild, with only 2D body joint or semantic segmentation maps ground truth. Our weakly-supervised model relies on all practically useful data in order to bootstrap a self-supervised system at later stages. Hence we do not discard 3D data when we have it, and aim to use it to circumvent the missing link: images in the wild with 3D pose and shape ground truth. 
Then, one can define weakly supervised losses for \emph{keypoint alignment:} $L_{KA} = \frac{1}{N_{j}} \sum_{i=1}^{N_j}\left\|\mathbf{J}_{i} - \widehat{\mathbf{J}}_{i}\right\|_{2}^{2}$, \emph{semantic body-part alignment:} $L_{BA}\left(\widehat{\mathbf{S}}, \mathbf{V}\right)$, and \emph{the prior:} $L_{\psi} = \psi_{nf}\left(\ff(\thetab)\right)$ (or $\psi_{nf}\left(\zz\right)$ when working in the latent space). The weakly supervised loss is a combination of multiple losses, plus a term that regularizes the shape parameters
\begin{equation}
L_{ws} = L_{KA} + L_{BA} + L_{\psi} + \left\lVert\betab\right\rVert_{2}^{2} \label{eq:ws_loss}
\end{equation}
The total loss will be $L_{total} = L_{fs} + L_{ws}$. For a graceful transition between supervision regimes, during fully supervised training we use $L_{total}$, then switch to $L_{ws}$ in the weakly supervised phase.

\section{Experiments}
\label{sec:exps}

\begin{figure*}[!htbp]
\begin{center}
        \includegraphics[width=0.85\linewidth]{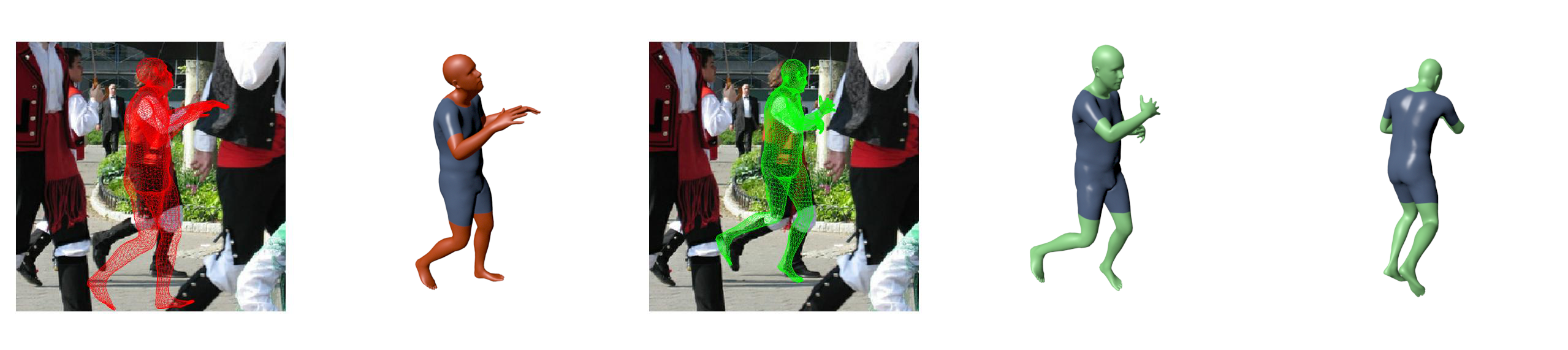}
        \includegraphics[width=0.85\linewidth]{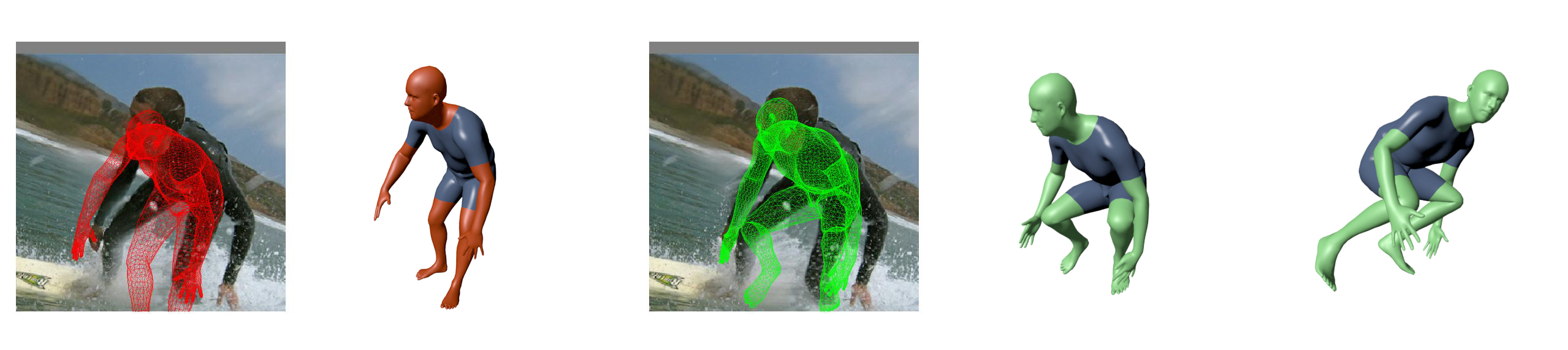}
        \includegraphics[width=0.85\linewidth]{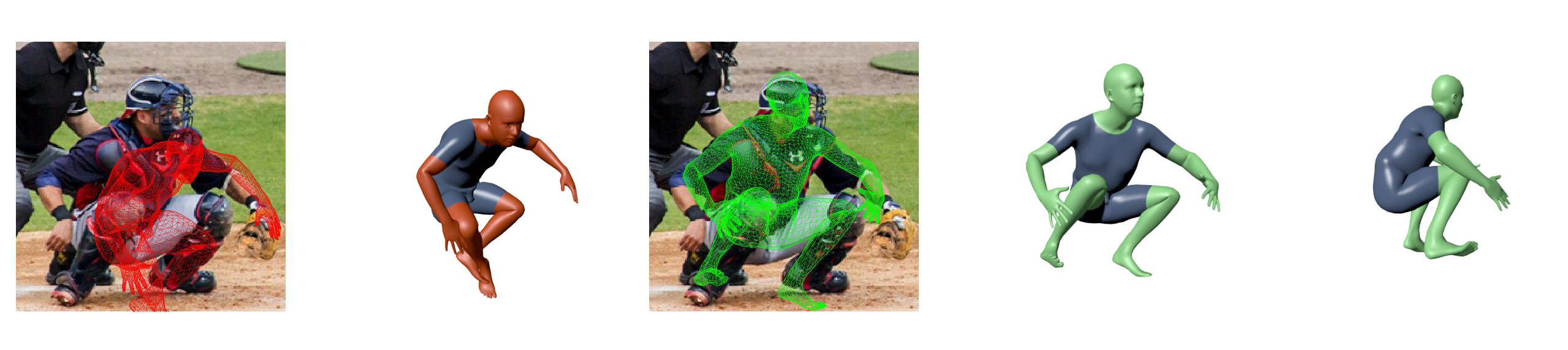}
        \includegraphics[width=0.85\linewidth]{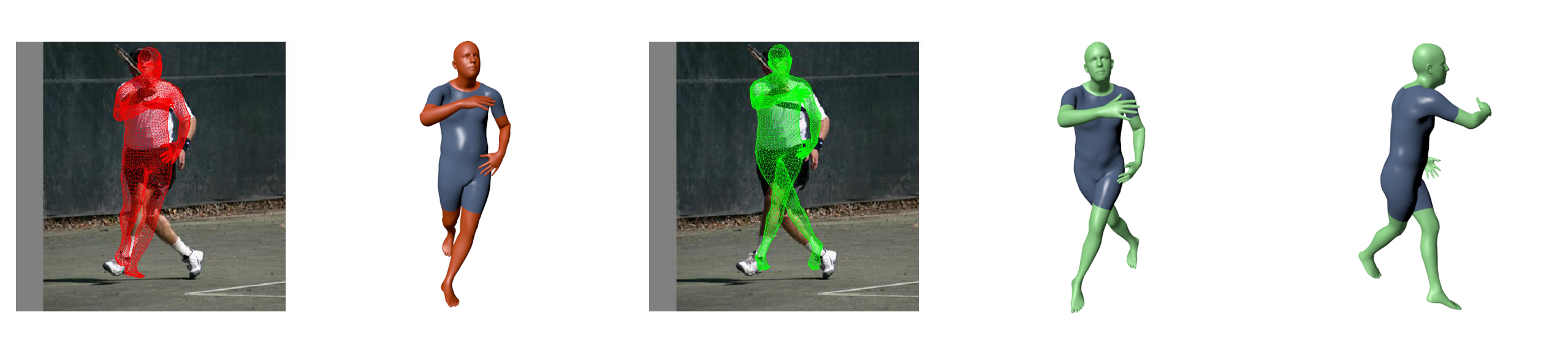}
        \includegraphics[width=0.85\linewidth]{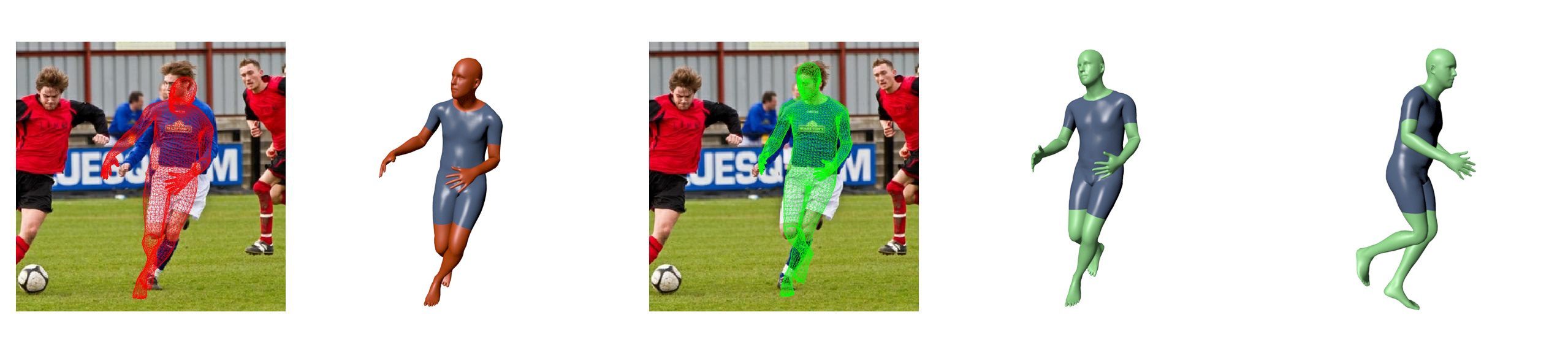}
        \includegraphics[width=0.85\linewidth]{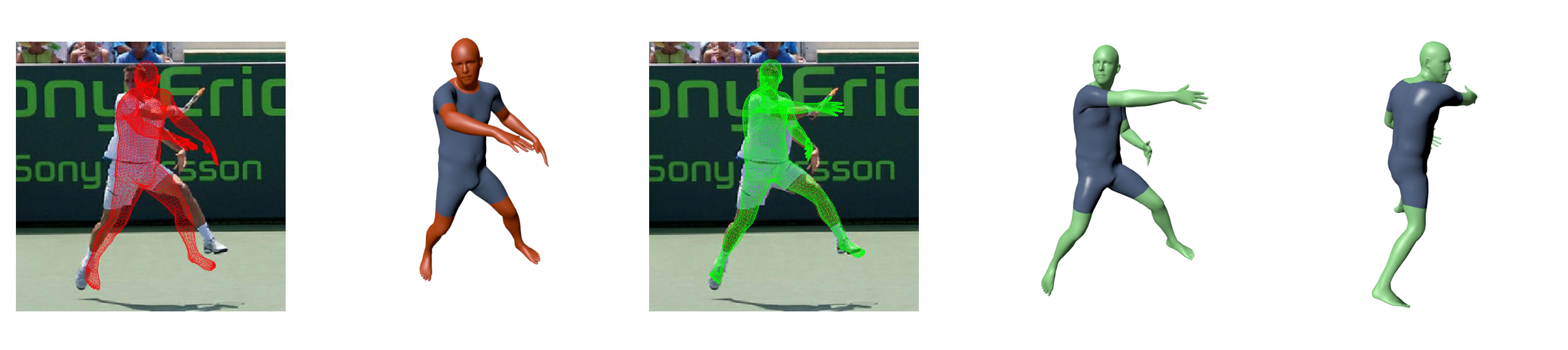}        
\end{center}
\caption{\textbf{Reconstruction results of models trained weakly-supervised using COCO (best seen in color)}. Starting from a network fully supervised on H80K (\textbf{red}), we fine-tune with a weakly-supervised loss (\textbf{green}) and a normalizing flow kinematic prior. Notice considerable improvement in both alignment and the perceptual 3D estimates. Last column shows a different view angle for the WS estimate.}
\label{fig:side-by-side}
\end{figure*}

\noindent{\bf Datasets.} We run our fully supervised experiments on the Human80K (H80K) -- a representative subset sampled from Human3.6M (H3.6M) \cite{Ionescu14pami}. We also use H80K in order to train pose priors and for optimization experiments. We report errors in the form of MPJPE (mean per-joint position error) and MPVPE (mean per-vertex position error) all in 3D.

We split the training set of H80K (composed of $\approx54,000$ images) into train, eval and test. As there are no publicly available statistical body model fittings for H80K data, we had to build them ourselves. Based on the ground truth 3D joint positions $\widehat{\mathbf{J}}_{3D}$ (this is used to retrieve pose $\widehat{\thetab}$) and the available 3D subject scans (used to retrieve shape $\widehat{\betab}$) provided with the dataset, we optimize a fitting objective (solved using BFGS). We then project the 3D meshes associated to motion captured body configurations in each frame, to obtain ground truth 2D annotations, $\widehat{\mathbf{J}}_{2d}$ and $\widehat{\mathbf{S}}_{2d}$. We thus have full supervision on H80K in the form of $\left(\widehat{\thetab}, \widehat{\betab}, \widehat{\mathbf{J}}_{2d}, \widehat{\mathbf{S}}_{2d} \right)$ for each image in the training set.

We also use CMU \cite{cmu2018} and AMASS \cite{mahmood2019amass} to train 3D pose priors. Both datasets have publicly available kinematic model fittings and we used the $\widehat{\thetab}$ values to train our normalizing pose model $p_{\bphi}(\thetab)$. This results in priors over ambient and latent spaces, $\psi_{nf}(\ff(\thetab))$ and $\psi_{nf}(\zz)$, respectively. Similar models were trained for H80K. %In our experiments, we also use the GMM prior $\psi_{gmm}(\thetab)$ of \cite{bogo2016}. 

For weakly supervised learning `in the wild', we use a subset of 15,000 images from COCO 2014 \cite{MsCOCO}. 
The dataset has no 3D ground truth, but offers 2D annotations for human body joints $\widehat{\mathbf{J}}_{2d}$, as well semantic segmentation of body parts $\widehat{\mathbf{S}}_{2d}$.  We split the data in 14,000 examples for training and 1,000 for testing, and use it for building the weakly supervised models. We refer to models trained using 2D body joints and semantic body part losses as KA and BA, respectively.

\noindent{\bf Optimization with Different Priors and Losses.} In order to analyze the impact of priors and semantic segmentation losses on optimization, we choose H80K where ground-truth is available for all components including 3D camera, pose and shape. We perform non-linear optimization with the objective function as defined in \eqref{eq:ws_loss}, where $L_{\psi}$ is changed to accommodate all the various priors, and $L_{KA}$ and $L_{BA}$ are studied both together and independently.

We evaluate different prior types: \textit{\bf i)} $\psi_{gmm}(\thetab)$ -- GMM \cite{bogo2016}, \textit{\bf ii)} $\psi_{nf}(\ff(\thetab))$ and $\psi_{nf}(\zz)$ -- normalizing flow in ambient and latent space, as given by \eqref{eq:nflow_orig}, and \eqref{eq:nflow_wrp}, using either the angle-axis or the 6D representation, \textit{\bf iii)}  $\psi_{VPoser}(\zz)$ -- the variational auto-encoder VPoser of \cite{pavlakoscvpr2019}, \textit{\bf iv)} $\psi_{hmr}(\thetab)$ -- the discriminator of \cite{Kanazawa2018}. 
 
We also evaluate different datasets (CMU, H80K, AMASS) for prior construction, different loss functions based on either body joints/keypoints or semantic segmentation of body parts (KA and BA). To directly compare with VPoser, we train a light-weight normalizing flow prior ($\approx 93,000$ parameters compared to $\approx 344,000$ for VPoser), with the same operating speed, and constructed on the same dataset (AMASS) and train/test splits. 

To isolate confounding factors, optimization is performed using the ground-truth 2D joints (KA) and body part labels (BA), under a perspective projection model, by using the loss defined at \eqref{eq:ws_loss}. Optimization relies on BFGS with analytical Jacobians, obtained through automatic differentiation. We start with four different initializations and report the solution with the smaller loss (N.B. this does not require observing the ground truth). We consider four different global rotations, and initialize parameters with $\mathbf{0}$, for pose (either in ambient or latent space) and shape.

We test the model on 500 images and report results in table \ref{tbl:fitting_experiment_h80k}. The best results are achieved by normalizing flow priors when optimization is performed in latent space. By using both keypoint and body part alignment-based self-supervision, the results improve. The 6D rotation representation has a slight edge over the angle-axis. The light-weight normalizing flow trained on AMASS is the best performer, surpassing VPoser even at a third of its capacity. Note that we do not have to balance two terms (the reconstruction loss and the KL-divergence), as normalizing flows support exact latent-variable inference. Additionally, VPoser requires posing meshes \emph{during training}, whereas normalizing flow models do not.
\begin{figure}
\capbtabbox{%
    \small
    \begin{tabular}[t]{|l|r|r|}
    \hline
    \textbf{Method}  & {\textbf{Error (cm)}}\\ 
    \textit{prior, dataset, representation, features} & {MPJPE/MPVPE}\\
    \hline
    \hline
    $\psi_{gmm}(\thetab)$, CMU, AA, KA &  $7.9/10.4 $\\
    $\psi_{gmm}(\thetab)$, CMU, AA, KA + BA & $6.9/9.6$\\
    \hline
    $\psi_{VPoser}(\zz)$, AMASS, 6D, KA &  $4.6/6.7 $\\
    $\psi_{nf}(\zz)$, AMASS, 6D, KA &  $\mathbf{4.3/6.0}$\\
    \hline
    $\psi_{hmr}(\thetab)$, H3.6M, RM, KA &  $11.9/15.3 $\\
    \hline
    $\psi_{nf}(\ff(\thetab))$, CMU, AA, KA &  $6.2/8.4$\\
    $\psi_{nf}(\ff(\thetab))$, CMU, AA,  KA + BA &  $6.0/8.1$\\
    \hline
    $\psi_{nf}(\zz)$, CMU, AA, KA &  $5.0/7.1$\\
    $\psi_{nf}(\zz)$, CMU, AA,  KA + BA &  $4.9/6.9$\\
    \hline
    $\psi_{nf}(\ff(\thetab))$, CMU, 6D, KA &  $6.1/8.4$\\
    $\psi_{nf}(\ff(\thetab))$, CMU, 6D, KA + BA & $5.8/8.0$\\
    \hline
    $\psi_{nf}(\zz)$, CMU, 6D, KA &  $5.1/6.8$\\
    $\psi_{nf}(\zz)$, CMU, 6D, KA + BA & $4.8/6.6$\\
    \hline
    \hline
    $\psi_{nf}(\ff(\thetab))$, H80K, AA, KA &  $5.4/7.5$\\
    \hline
    $\psi_{nf}(\zz)$, H80K, AA, KA &  $4.4/6.1$\\
    \hline
    \end{tabular}
}
{
\caption{\small Optimization-based pose and shape estimation experiments with evaluation on the ground truth of H80K dataset. Priors are learned on the training sets of CMU, AMASS or H80K. The HMR discriminator has the largest errors, as it was arguably designed for use with deep neural network losses, and not for model fitting. Optimizing in latent space (using normalizing flows) and semantic alignment always helps. The 6D representation performs slightly better than angle-axis. The best performers are objective functions that include normalizing flow priors trained on H80K or AMASS. VPoser performs slightly worse than our normalizing flow prior, even though it also encodes and decodes 6D rotations. \textit{Notation:} AA = angle-axis representation, 6D = 6 dimensions rotation representation, RM = rotation matrices, KA = keypoint alignment, BA = body alignment.}
\label{tbl:fitting_experiment_h80k}
}
\end{figure}

\begin{table*}[!htbp]
    \small
    \centering
    \begin{tabular}[t]{|l|r|r|r|r|r|r|}
    \hline
    \textbf{Percentage Supervised}  & {0\%} & {20\%} & {40\%} & {60\%} & {80\%} & {100\%}  \\ 
    \hline
    \hline
    \textbf{FS (mm)} & $649/677$ & $117/136$ & $101/118$ & $93/109$ & $86/102$ & $83/97.15$\\
    \hline
    \textbf{WS (mm)} & $\mathbf{123/140}$ & $\mathbf{97/111}$ & $\textbf{92/108}$ & $\mathbf{90/106}$ & $\mathbf{85/101}$ & $84/98.85$ \\
    \hline
    \end{tabular}
    \caption{\small Ablations on H80K, reported as MPJPE/MPVPE metrics in millimeters. Notice the impact of weakly supervised losses (WS), especially in the fully supervised (FS) regime with small training sets, as well as for the model initialized randomly (column two, $0\%$ supervision). }
\label{tbl:ablation_fs_ws_h80k}
\end{table*}

\noindent{\bf Fully to Weakly Supervised Transfer Learning.} 
We present experiments and ablation studies showing how the weakly supervised training of shape and pose parameters $\left(\thetab, \betab\right)$ can be successful in conjunction with the proposed normalizing flow priors and self-supervised losses.

For this study, we split H80K into two parts where we keep 5 subjects for training (S1, S5, S6, S7 and S8) and two subjects (S9, S11) for testing.

We further split the training set into partitions of 20\%, 40\%, 60\%, 80\%, 100\%. We initially train the network fully supervised (FS) on the specific partition of the data using $L_{fs}$ loss. We train the fully supervised model for 30 epochs, then continue in a weakly supervised (WS) regime based on $L_{ws}$ on all the data. 
In table \ref{tbl:ablation_fs_ws_h80k} we report MPJPE/MPVPE for the ablation study. Notice  that in all cases weak supervision improves performance whenever additional image data is available.

We also check that our methodology compares favorably to a similar method HMR \cite{Kanazawa2018} which we retrained on H80K. In this case our model achieves 84mm MPJE whereas HMR has 88mm.\footnote{Based on HMR's Github repository, we identify a total of $\approx$27M trainable parameters. 
Our model has 6 stages, each with $5 \times 7 \times 7 \times 128 \times 128$ parameters resulting in $\approx$24M trainable parameters. } We were not able to train on their split and retargeting of H3.6M, as their training data was not available.

\begin{figure}[!htbp]
\begin{center}
    \includegraphics[width=0.49\linewidth]{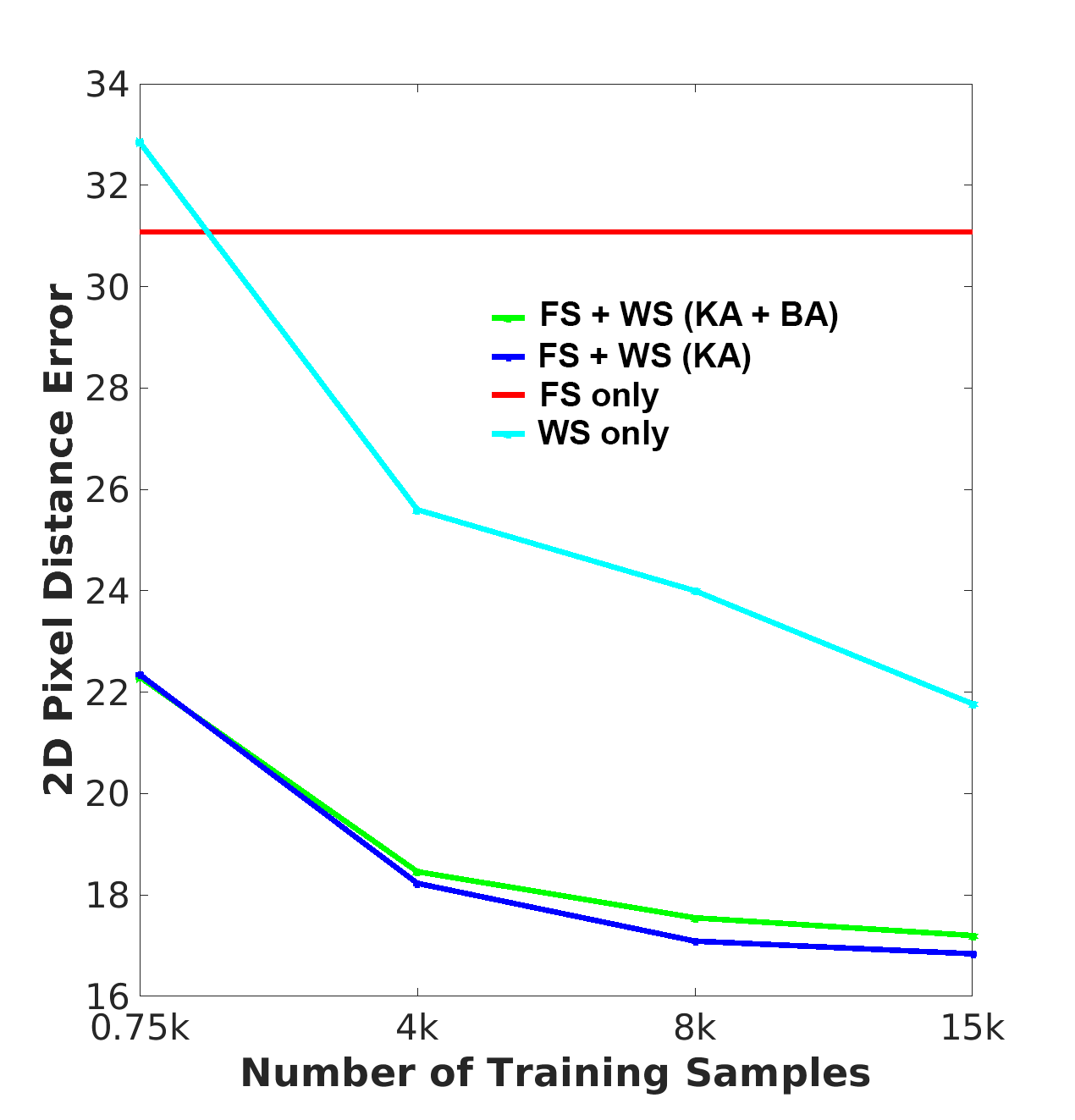}
    \includegraphics[width=0.49\linewidth]{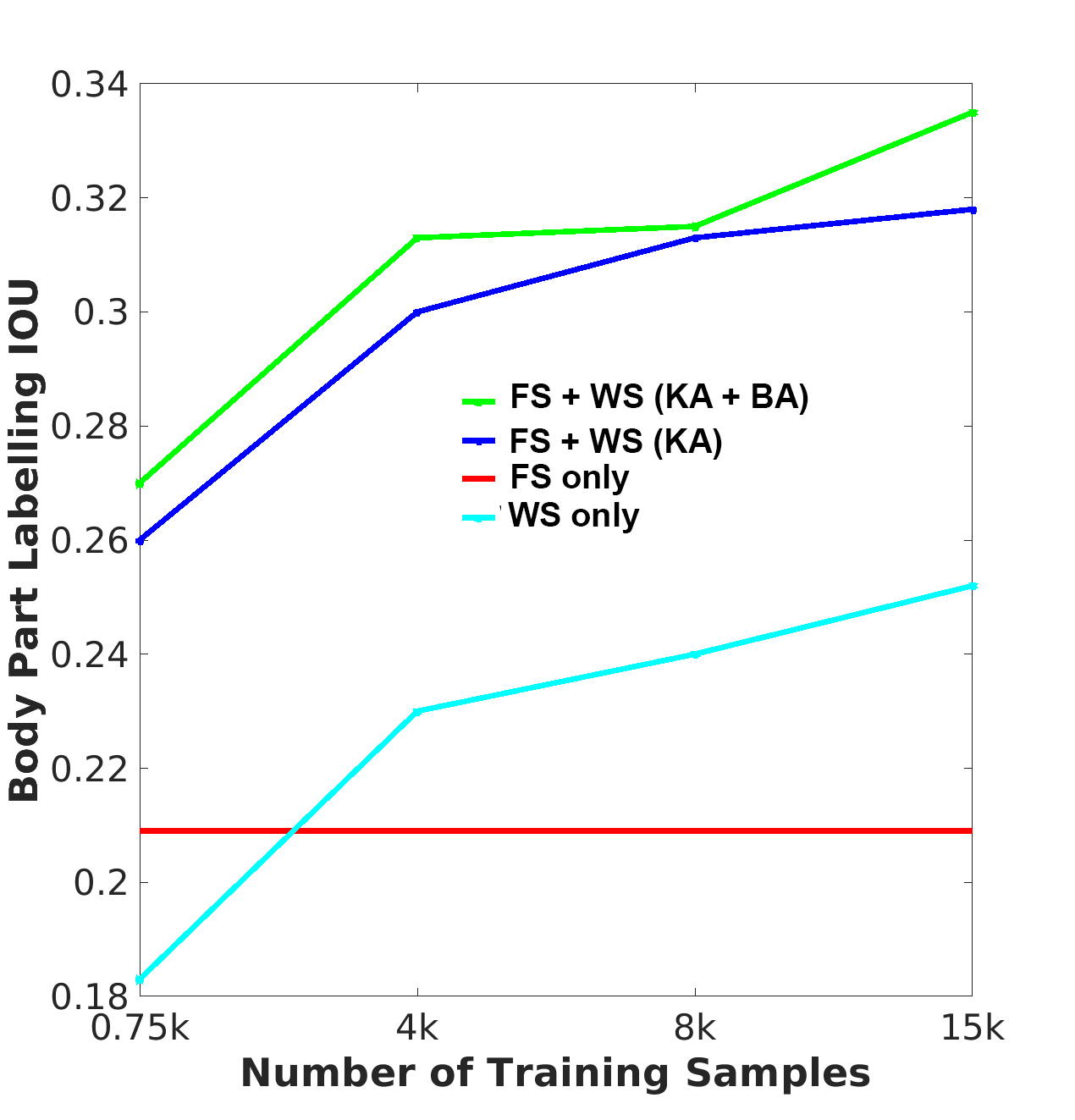}
\end{center}

\caption{\small \textit{Left and Right}: Weakly supervised experiments on COCO with different loss combinations (KA, KA+BA) and different amounts of training data. The baseline is obtained by running the network trained fully supervised on H80K. WS Only is trained only on COCO.}
\label{fig:ka-kb-weight_sensitivity-figure}
\end{figure}

\noindent{\bf Weakly Supervised Transfer for Images in the Wild.}  In order to validate our network predictions beyond a motion capture laboratory, `in the wild', we refined the network on the subset of COCO which has body part labelling available. We started with a network pre-trained on H80K, then continued training on COCO using the complete loss. As ground truth 3D is not available for COCO, we monitor errors between ground truth and estimated 2D projections of the 3D model joints, and the IoU semantic body part alignment metrics. As shown in fig. \ref{fig:ka-kb-weight_sensitivity-figure}, in all cases the pixel error of the projected 2D joints decreased consistently, as a result of weakly supervised fine tuning. A similar trend can be seen for the IoU metric computed for body part alignment, illustrating the importance of a segmentation loss. We explicitly run two configurations, one in which we only use the keypoints alignment (KA) and another based on body part alignment (KA+BA). 
\begin{table}[!htb]
    \begin{tabular}[t]{|c|l|r|r|}
    \hline
    & Method & MPJPE (mm) & MPJPE-PA (mm) \\
    \hline
    \multirow{8}{*}{\rotatebox[origin=c]{90}{\textbf{STATIC}}} & HMR \cite{Kanazawa2018} &- &$81.3$ \\
    \hhline{~---}
    & Kanazawa et al. \cite{humanMotionKanazawa19} &- &$72.6$ \\
    \hhline{~---}
    & SPIN \cite{kolotouros2019learning} (static fits) &- &$66.3$ \\
    \hhline{~---}
    & SPIN \cite{kolotouros2019learning} (best) &- &$59.2$ \\
    \hhline{~---}
    & FS & $95$ & $61.3$ \\
    \hhline{~---}
    & FS+OPT (KA) & $95$ & $60.3$ \\
    \hhline{~---}
    & FS+OPT (KA+BA) & $91.4$ & $58.87$ \\    
    \hhline{~---}
    & FS+WS (KA+BA) & $\mathbf{90.0}$ & $\mathbf{57.1}$ \\ 
    \hline
    \hline
    \multirow{5}{*}{\rotatebox[origin=c]{90}{\textbf{VIDEO}}} & VIBE \cite{kocabas2019vibe}(16 frames) &$82.9$ &$51.9$ \\
    \hhline{~---}
    & FS+OPT(KA+BA+S, 16 frames) &$82.8$ &$52.2$ \\
    \hhline{~---}
    & FS+WS+OPT(KA+BA+S, 4 frames) &$84.5$ &$54.5$ \\
    \hhline{~---}
    & FS+WS+OPT(KA+BA+S, 8 frames) &$82.0$ &$51.4$ \\
    \hhline{~---}
    & FS+WS+OPT(KA+BA+S, 16 frames) &$\mathbf{80.2}$ &$\mathbf{49.8}$ \\
    \hline
    \end{tabular}

\caption{\small Results on the 3DPW test set for two regimes: \textbf{static} and \textbf{video}. FS is fully supervised, FS+OPT are predictions from FS with optimization. FS+WS are results for self-supervised refinement of the FS model on MS COCO. ‘S’ stands for smoothing in the video regime. MPJPE is the mean per joint position error, whereas MPJPE-PA is the error after Procrustes alignment. \textbf{Static}: we observe that the self-supervised training did not affect the performance of the 3D predictions. The semantic alignment loss reduces error more than only keypoints alignment. Perceptually, image alignment is also much better for BA than KA, even
when it does not immediately produce significant 3D quantitative improvements. \textbf{Video}: the best performer is our FS+WS (KA+BA) model, further optimized over 16 frames with the temporal smoothing term.
}
\label{tbl:3dpw_experiments}
\end{table}

A potentially interesting question is whether the 3D prediction is affected by a self-supervised refinement. We run experiments on 3DPW \cite{vonMarcard2018} which consists of $\approx 60,000$ images containing one or more humans performing various actions in the wild. The subjects were recorded using IMUs so shape and pose parameters were recovered. We used the training data as supervision, and evaluate on the test set. We report results for a model trained only with full supervision, as well as results of refining the fully supervised (feed-forward) estimate by further optimizing the KA, and KA+BA losses against the predicted 2D outputs (keypoint and body part alignment). After training the network in the weakly supervised regime we obtain better accuracy, showing that 3D prediction quality is preserved. We show the results in table \ref{tbl:3dpw_experiments}. To the best of our knowledge, these are the lowest errors reported so far on the 3DPW test set in a static setting.

\noindent{\bf Temporal optimization.} We also experiment in the temporal setting, on batches of 4, 8 and 16 consecutive frames drawn from the 3DPW dataset. Starting from the best results obtained per frame in the static setting, we do a whole batch optimization. Different from the $L_{ws}$ objective, now the shape parameters are tied across frames, with an additional term that enforces smoothness between adjacent temporal pose parameters (in latent space):
\begin{align}
    L_{smooth} = \sum_{t=2}^{N_f} \left\|\zz^{t} -\zz^{t-1}\right\|_{2}^{2}
\end{align}

The weight for this term is set to be $50\times$ the weight of the prior, as we expect a lower variance for pose dynamics. We compare our method with the recent work of \cite{kocabas2019vibe}, showing the results in table \ref{tbl:3dpw_experiments}. As in the static setting, these are also the lowest errors reported so far.

\section{Conclusions}

We have presented large scale weakly supervised deep learning-based models for 3D human pose and shape estimation from monocular images and video. Key to scalability is unlocking the ability to exploit human statistics implicitly available in large, diverse image repositories, which however do not come with detailed 3D pose or shape supervision. Key to making such approaches feasible, in terms of identifying model parameters with good generalization performance, is the ability to design training losses that are tightly controlled by both the existing prior knowledge on human pose and shape, and by the image and video evidence. 

We introduce latent normalizing flow representations and dynamical models, as well as fully differentiable, structured, semantic body party alignment (re-projection) loss functions which provide informative feedback for self-supervised learning. In extensive, large-scale experiments, using both motion capture datasets like CMU, Human3.6M, AMASS, or 3DPW, as well as `in the wild' repositories like COCO, we show that our proposed methodology achieves state-of-the-art results in both images and video, supporting the claim that constructing accurate models based on large-scale weak supervision `in the wild' is possible.

\clearpage
\section{Appendix}
\subsection{Architectures}
We describe the architectures used to construct a normalizing flow prior. We illustrate the case where the input is represented in terms of 6D rotation variables. We assume $23$ joints (corresponding to the SMPL kinematic hierarchy), each with $6$ dimensions representing each rotation. Hence, the total dimension of the body pose representation is $138$. For other rotation representations (e.g. angle-axis, rotation matrix, etc.) the same procedure applies.

\paragraph{Low-capacity version} We test a low-capacity normalizing flow architecture with the following structure: FC138-PReLU-FC138-PreLU-FC138-PreLU-FC138-PReLU-FC138, with a total of $95,914$ trainable parameters. In comparison, VPoser\cite{pavlakoscvpr2019} uses $344,190$ parameters. Note that in the backward pass from latent to ambient space we do not use matrix inversions -- the fully connected layers are applied in a standard way.

\paragraph{Real-NVP version} We also use a more complex normalizing flow architecture, which replaces the PreLU activation unit with a Real-NVP step. The structure is then FC138-RNVP-FC138-RNVP-FC138-RNVP-FC138-RNVP-FC138-RNVP-FC138, with a total of $331,462$ trainable parameters. For the Real-NVP unit, we use a simple FC128-Tanh-FC128-Tanh-FC69 architecture. 

\paragraph{Training} We use a custom TensorFlow implementation for all architectures. In training, the batch size is set to $64$, and we use  ADAM optimization with an initial learning rate of $1e-4$ and an exponential decay rate of $0.99$ at every $10,000$ steps. The training is stopped after $200,000$ steps. For the AMASS dataset, this corresponds to $\approx 4$ epochs.

\subsection{Translation Estimation from 2d Keypoints}

For all of our experiments, we assume a perspective camera model. In this case, one unknown is the global model translation, $\mathbf{T}$, which has to be either predicted or estimated.
Unfortunately, predicting a 3d translation directly is difficult with neural networks. In order to circumvent this, we propose the following solution: given a posed mesh $\mathbf{M}(\thetab, \betab)$, with skeleton joints $\mathbf{J}_{3d} = \{ \mathbf{J}^{3d}_{i}, i=1 \ldots N_{j} \}$, projected skeleton joints $\mathbf{J}_{2d} = \{ \mathbf{J}_{i}, i=1 \ldots N_{j} \}$ and detected 2d joint locations $\{\widehat{\mathbf{J}_i}\}$, we rewrite the keypoint alignment error as:
\begin{align}
    L_{KA} &=\frac{1}{N_j}\sum_i \|\mathbf{J}_i - \widehat{\mathbf{J}_i}\|_2 \nonumber\\
           &= \frac{1}{N_j}\sum_i \|\Pi(\mathbf{J}^{3d}_i + \mathbf{T}) - \widehat{\mathbf{J}_i}\|_2
    \label{eq:translation_estimation}
\end{align}

where $\Pi$ is the perspective projection operator. By relaxing the operator to a weak-perspective one, $\Pi_W$, we can solve for  translation directly, by using least-squares:
\begin{align}
    \mathbf{T}^{*} = \argmin_{\mathbf{T}} \frac{1}{N_j}\sum_i \|\Pi_W(\mathbf{J}^{3d}_i + \mathbf{T}) - \widehat{\mathbf{J}_i}\|^{2}_{2}
    \label{eq:translation_lsqr}
\end{align}

Note that \eqref{eq:translation_lsqr} is used only to predict the global translation, whereas \eqref{eq:translation_estimation} is used afterwards to compute the keypoint alignment loss, based on the estimated $\mathbf{T}^*$. Gradients will flow to all the variables of the network, through both layers implementing the above operations.

\subsection{Normalizing Flows and VPoser on 3DPW}

\begin{figure}[!ht]
\begin{center}
    \includegraphics[width=1.\linewidth]{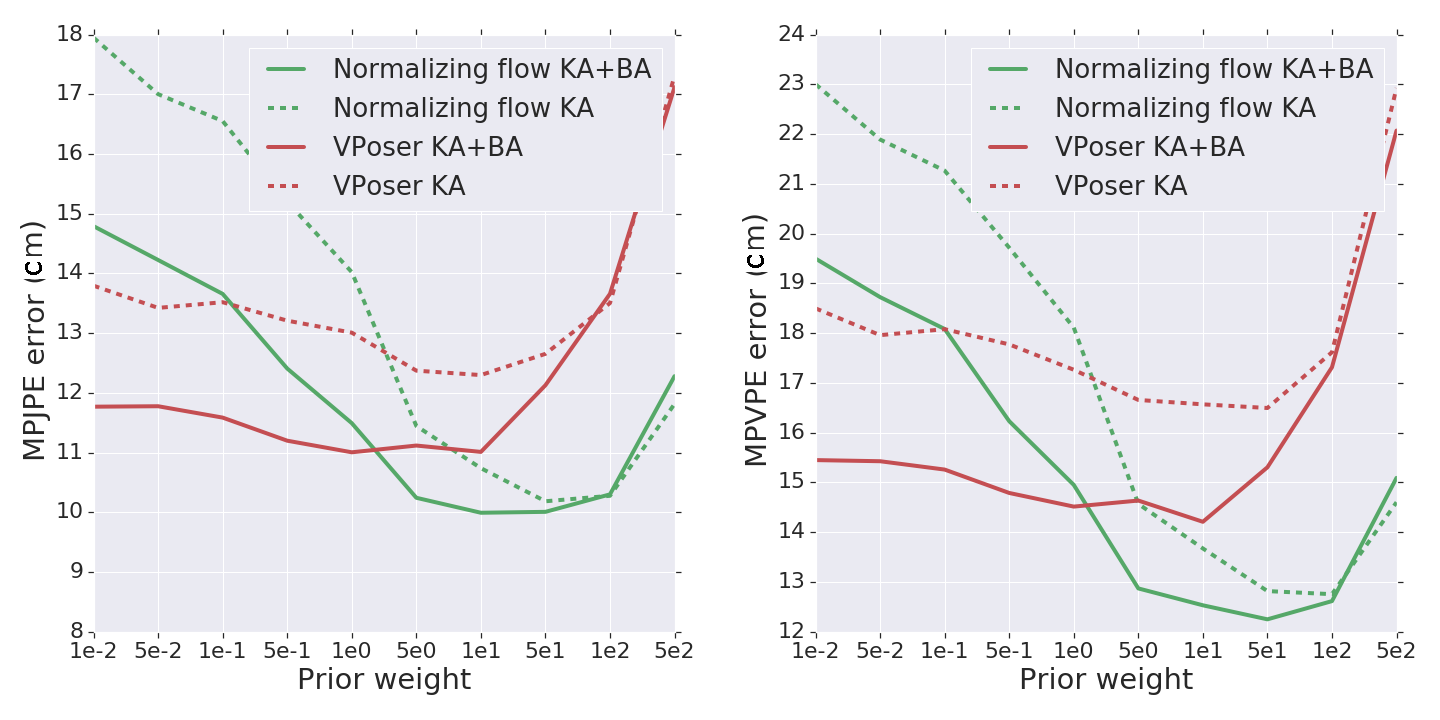}
\end{center}
\caption{Reconstruction errors, MPJPE (left plot) and MPVPE (right), for different priors and loss functions. Notice that normalizing flow priors reduce the reconstruction error in all cases.}
\label{fig:3dpw_fitting}
\end{figure}

In this experiment, we compare our light-version normalizing flow prior (trained on AMASS) with the prior of \cite{pavlakoscvpr2019}, on 500 random images sampled from the 3DPW dataset. In this study, the 2d keypoints and semantic segmentation are predictions from a deep-neural network we trained, and images have ground-truth 3d meshes which permits evaluation. We fit the SMPL model in the same conditions (starting from 4 globally rotated 0-mean latent space kinematic initializations, using both KA and KA+BA losses), for both priors, and report errors in fig. \ref{fig:3dpw_fitting}.

\subsection{Self-supervised Learning on\\COCO and OpenImages}

In order to further explore the effect of additional self-supervision to our training process, we extended the set of in-the-wild images with a subset of OpenImages\cite{OpenImages}. OpenImages contains various annotations from which we used the ones related to people (bounding boxes) in various shapes and poses. We kept the images on which the keypoint detector component of our network predicted enough keypoints with high confidence. Using the 2D keypoints and the semantic segmentation predictions from the network we extended the training data with up to 70,000 samples, including the initial COCO data mentioned in the paper. We gradually increased the amount of data from 10\% up to 100\% used to further train the network and we show results on the 3DPW test set. As can be seen in table 1, the 2D joint error is decreasing and the overlap mIOU metric is increasing showing that with more self supervision the predictions of the network get better. As can be seen in the paper this also leads to better 3D predictions.

\begin{table}[!htbp]
    \begin{tabular}[t]{|l|r|r|}
    \hline
    Method & 2D Joints Error (pixels) & mIOU \\
    \hline
    FS & $9.13$ & $42.5$ \\
    \hline    
    WS+KA+BA-10\% & $7.2$ & $45.5$ \\    
    \hline
    WS+KA+BA-30\% & $6.27$ & $47.86$ \\    
    \hline
    WS+KA+BA-60\% & $\mathbf{5.4}$ & $47.06$ \\    
    \hline
    WS+KA+BA-100\% & $5.8$ & $\mathbf{50.0}$ \\
    \hline
    WS+KA-10\% & $6.91$ & $41.8$ \\
    \hline
    WS+KA-30\% & $6.0$ & $42.2$ \\  
    \hline
    WS+KA-60\% & $5.7$ & $43.0$ \\
    \hline
    WS+KA-100\% & $5.6$ & $44.0$ \\
    \hline
    \end{tabular}
{
\caption{\small Self supervised experiments on the 3DPW test set using COCO and OpenImages data for additional training. FS identifies the model trained only using full supervision. WS is the model trained weakly supervised. KA and BA denote the keypoint, respectively the body part alignment losses. We gradually increased the amount of self supervised data used for refining the network, which was initially trained fully supervised. We observe that the 2D joint error (measured in pixels) is decreasing as we add more data. As expected, the mIOU metric is increasing--more so in the case where the KA+BA loss is used denoting better alignment. Usually decreases in KA correlate with increases in BA, although this is not always the case-- one can expect that a certain lack of calibration between the 2d detected skeletons and the 3d SMPL counterpart, or an aggressive maximization of overlap when clothing makes it difficult to correctly segment body parts, could lead to potential inconsistencies between the trends of the two losses. In practice we find the model image alignment to be much better for BA than for KA, with 3d reconstructions that are perceptually good. 
}
}
\label{tbl:ssupervised-experiments2}
\end{table}
\bibliographystyle{splncs04}
\bibliography{egbib}
\end{document}